\documentclass[runningheads]{llncs}
\usepackage{graphicx} 
\usepackage{subfig}
\begin{document}
\title{Reinforcement Learning with Attention that Works: A Self-Supervised Approach }
\titlerunning{Reinforcement Learning with Attention that Works}
%
\author{Anthony Manchin \and
Ehsan Abbasnejad \and
Anton van den Hengel}
\authorrunning{A. Manchin et al.}
%
\institute{The Australian Institue for Machine Learning - The University of Adelaide
\email{\{anthony.manchin, ehsan.abbasnejad, anton.vandenhengel\}@adelaide.edu.au}}

\maketitle              
\begin{abstract}
Attention models have had a significant positive impact on deep learning across a range of tasks. However previous attempts at integrating attention with reinforcement learning have failed to produce significant improvements. We propose the first combination of self attention and reinforcement learning that is capable of producing significant improvements, including new state of the art results in the Arcade Learning Environment. Unlike the selective attention models used in previous attempts, which constrain the attention via preconceived notions of importance, our implementation utilises the Markovian properties inherent in the state input. Our method produces a faithful visualisation of the policy, focusing on the behaviour of the agent. Our experiments demonstrate that the trained policies use multiple simultaneous foci of attention, and are able to modulate attention over time to deal with situations of partial observability.

\end{abstract}
\section{Introduction}

Research on reinforcement learning (RL) has seen accelerating advances in the past decade. In particular, methods for deep RL have made tremendous progress since the seminal work of Mnih \emph{et al.}~\cite {mnih2015humanlevel} on Deep Q-Networks. A number of different approaches have progressed the state of the art on simulated tasks - in particular in the Arcade Learning Environment \cite{bellemare13arcade} - to, or above that of human players. The major focus of attention of many of these methods is the learning and optimization of the policies. However, one thing that all of these approaches have in common is that they process the raw input data through a convolutional neural network (CNN). 

Regardless of the chosen policy optimisation method, the underlying neural network is responsible for interpreting and encoding useful representations of spatial and temporal information from the input state. While significant efforts have focused on methods for policy optimisation, the techniques for encoding the observations have received relatively less attention. While many methods thus use generic, off-the-shelf CNN architectures, we instead focus on this as the main subject of study.

Taking inspiration from other areas of deep learning, we investigate the benefits of incorporating self-attention into the underlying network architecture. Attention models were applied with remarkable success to complex visual tasks such as video and scene understanding~\cite{Han:2018:EMR:3265987.3265996,DBLP:journals/corr/abs-1812-01855,Fang:2018:ALE:3240508.3240571}, natural language understanding including machine translation~\cite{DBLP:journals/corr/BahdanauCB14,AAAI1816534}, and generative models using generative adversarial networks (GANs)~\cite{DBLP:journals/corr/abs-1711-10485,DBLP:journals/corr/abs-1802-09070}. We naturally assume that attention can similarly benefit RL tasks. Although previous attempts to integrate attention with RL have been made, these attempts have largely used hand-crafted features as inputs to the attention model \cite{DBLP:journals/corr/abs-1811-04407,DBLP:journals/corr/abs-1806-03960}. However the spatial and temporal information contained within the state input (when using stacked frames), is sufficient to satisfy the Markov assumption. This assumption states that the next state is entirely dependent upon the current state and is independent of previous states. This is important as RL operates entirely on this principle. As such, attention with respect to spatial and temporal reasoning would intuitively be a good starting point. 

For this reason we observe the work from Wang \emph{et al.}~\cite{DBLP:journals/corr/abs-1711-07971}, and their goal of improved attention through space, time and, space-time by combining self-attention with non-local filtering methods. The benefit of self-attention is the ability to compute representations of an input sequence by relating different positions of the input sequence. Their implementation achieves state of the art results on the Kinetics \cite{DBLP:journals/corr/KayCSZHVVGBNSZ17} and Charades \cite{DBLP:journals/corr/SigurdssonVWFLG16} datasets. However, the datasets they considered are large-scale video classification problems where the  changes in the input from time to time are minimal. In addition, the neural network is in a passive environment that does not require interaction. None the less, these challenges require spatial and temporal reasoning abilities that would be very useful for a reinforcement learning agent to posses. Taking inspiration from this work, we capitalise on the Markovian principle of the state input and propose a novel implementation of self-attention within the classical convolutional neural network architecture, as used by Mnih \emph{et al.}~\cite{mnih2015humanlevel}. In contrast to previous approaches, our self-attention approach utilises the Markovian properties of the state input. The contributions of our paper are as follows.

\begin{itemize}
	
	\item We provide a spatio-temporal self-attention mechanism for reinforcement learning and demonstrate that the network architecture has significant benefits in learning a good policy 
	 
	\item We present state-of-the-art results in the Arcade Learning Environment \cite{bellemare13arcade}. In particular, our approach significantly outperforms the baseline across a number of environments where the agent has to attend to multiple opponents and anticipate their movements in time.

 	\item We provide a visualisation of our approach that sheds new light on the policy learnt by the agent when equipped with the improved spatio-temporal representation, showing evidence of temporal attention.

\end{itemize}

\section{Related Work}

\subsubsection{Rl in video games}
Current state-of-the-art approaches for RL in the Arcade Learning Environment are built on top of the original network architecture proposed by Mnih \emph{et al.}~\cite{mnih2015humanlevel}. Alterations to this underlying network architecture have often included the implementation of recurrent neural networks (RNN). Hausknecht \emph{et al.}~\cite{DBLP:journals/corr/HausknechtS15} proposed replacing the fully connected layer, following the output of the last convolutional layer of the network with an LSTM. This allowed for a single frame input to be used, as opposed to sequentially stacked frames, with the LSTM integrating temporal information. Oh \emph{et al.}~\cite{DBLP:journals/corr/OhCSL16} also proposed using recurrent networks with their Recurrent Memory Q-Network (FRMQN). This memory-based approach used a mechanism based on soft attention to help read from memory and was evaluated with respect to solving mazes in Minecraft (a flexible 3D world).

A different approach by Fortunato \emph{et al.}~\cite{DBLP:journals/corr/FortunatoAPMOGM17} proposed adding parametric noise to the networks weights to aid efficient exploration, replacing conventional exploration heuristics. This modification  generally resulted in positive improvements, lending support to the idea that carefully considered improvements to the underlying network architecture can be beneficial.

In comparison to Hausknecht \emph{et al.} and Oh \emph{et al.} we utilise sequentially stacked frames as input, and augment the network with an attention model which is able to demonstrate improved temporal reasoning.

\subsubsection{Attention in RL}

Sorokin \emph{et al.}~\cite{DBLP:journals/corr/SorokinSPFI15}, which studied the effects of adding both ‘soft’ and ‘hard’ attention models to the network used by Mnih \emph{et al.}~\cite{mnih2015humanlevel} . These attention models received spatial information from a single frame processed by a CNN and temporal information from a RNN. Although this approach indicated some potential performance improvements under certain conditions, experiments were limited with results showing no systematic performance increases. In contrast, our work explores a different form of self-attention, and we demonstrate significant benefits, in both performance and interpretability for the resulting policy

Choi \emph{et al.}~\cite{DBLP:journals/corr/abs-1712-04603} also proposed combining attention with reinforcement learning for navigation purposes. This approach employed a Multi-focus Attention Network which used multiple parallel attention modules. This worked by segmenting the input, with each parallel attention layer attending to a different segment. This method was evaluated in a custom, synthetic grid-world environment, in which they reported better sample efficiency, in comparison to the standard DQN. In contrast our approach does not segment the input, instead using the output of the first convolutional layer as the input to the attention model.

Zhang \emph{et al.}~\cite{DBLP:journals/corr/abs-1806-03960} proposed an attention-guided imitation learning framework. They trained a model to replicate human attention with supervised gaze heatmaps. The input state was then augmented with this additional information. This style of attention fundamentally differs from that used in our work as it incorporates hand crafted features as input.
Gregor \emph{et al.}~\cite{inproceedings} also investigated visual attention using a glimpse sensory apporach. However this approach only investigated the integration of visual attention at the input layer, providing the network with different 'glimpses' of the full state.

More recently Yuezhang \emph{et al.}~\cite{DBLP:journals/corr/abs-1811-04407} proposed a model based upon the Broadbent filter model \cite{Broadbent_Filter}. This approach uses the optical flow calculated between two frames to construct an attention map, which was then combined with the output of the last convolutional layer in their network. This combination led to improved results when tested on a modified version of the toy problem  'Catch', originally inspired by Mnih \emph{et al.}~\cite{DBLP:journals/corr/MnihHGK14}. However the model was unable to replicate the same types of improvements in more visually complex domains.

\subsubsection{Visualising RL policies}
In recent attempts to better understand deep RL, two different visualisation techniques have been proposed. These techniques, although fundamentally different, both focus on revealing the relevant information used by the agent with respect to decision making. The first is a perturbation method, proposed by Greydanus \emph{et al.}~\cite{DBLP:journals/corr/abs-1711-00138}, which is a clear improvement over previous jacobian style visualisation methods \cite{Simonyan2013DeepIC}. Secondly, as shown by Weitkamp \emph{et al.}~\cite{DBLP:journals/corr/abs-1902-0056e} GRAD-cam visualisation techniques, as proposed by Selvaraju \emph{et al.}~\cite{DBLP:journals/corr/SelvarajuDVCPB16} can be used effectively with reinforcement learning. 

The network architectures, used by Greydanus \emph{et al.} and Weitkamp \emph{et al.} utilise a single frame input, with a RNN for temporal reasoning. This makes it difficult to visualize attention over the temporal dimension. Also Greydanus \emph{et al.}~\cite{DBLP:journals/corr/abs-1711-00138} clearly shows that their implementation of A3C is unable to attend to important information crucial for success in some environments. In contrast to this we clearly demonstrate an ability to visualise temporal reasoning. Another important point is that both papers display their visualisations over full scale, coloured frames. Although this is visually pleasing to the reader it is not representative of the input received by the agent, and as such may not reveal useful information to the research or reader. Instead we present our visualisations over the most recent frame in the time series, as received by the agent. This allows us to make keen observations as to why some environments are still prove challenging to current RL methods.


\section{Proposed Approach}

\begin{figure*}
	\includegraphics[width=\linewidth]{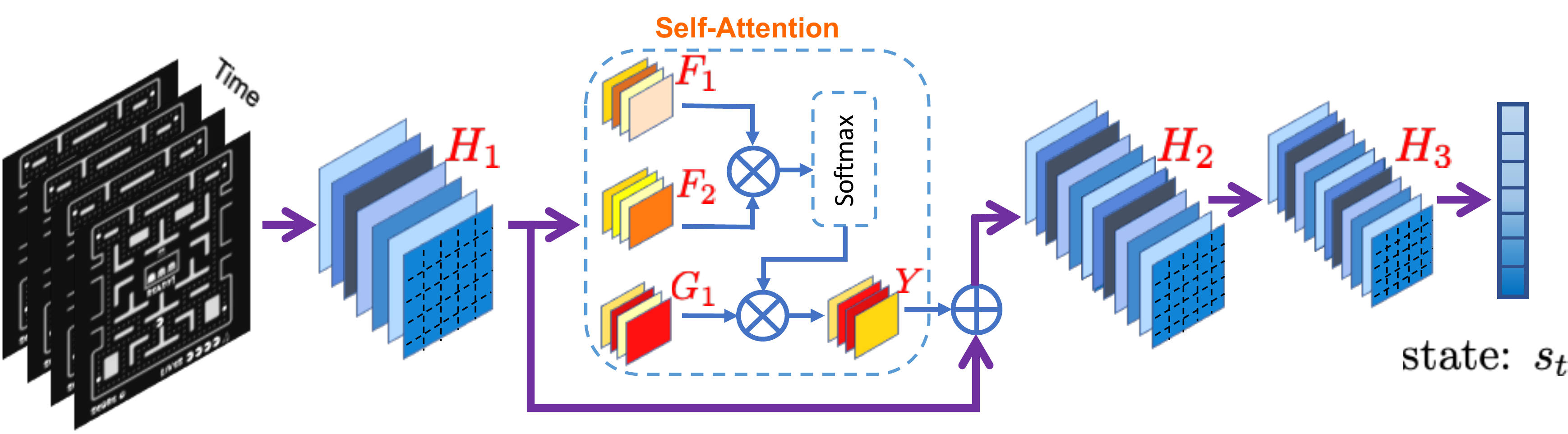}
	\caption{Overview of the proposed architecture. We introduce a self-attention module within the CNN used to process the input observations. The resulting policy benefits significantly from the capability of selective attention over space and time.}
	\label{fig:Self attending network (SAN) architecture}
\end{figure*}

We build upon the common practice of using a CNN to encode input observations into a state representation, suitable for complex decision making. Our main contribution is to incorporate a self-attention mechanism over space and time. This architecture will be shown to provide a significant benefit in learning effective policies. Previous works have attempted to provide a form of temporal attention by pairing CNNs with RNNs~\cite{DBLP:journals/corr/HausknechtS15}, but this resulted in limited success for temporal reasoning. There is additional evidence to suggest that agents acting on the simplified input of a single frame, may suffer difficulties in learning useful relationships over these inputs~\cite{DBLP:journals/corr/abs-1711-00138,DBLP:journals/corr/HausknechtS15}. We hypothesize that utilising relationships over parts of the input observations --~over space and time~-- are crucial for an agent to execute effective policies. This motivates the use of explicit attention mechanisms.

\subsection{Self-Attention Mechanism}

We specifically describe six instantiations of our general approach. Owing to the empirical nature of current research in deep learning, we conducted a thorough exploration of possible implementations of self-attention. 
Different domains have previously shown to be better addressed with different, sometimes conflicting implementation choices~\cite{DBLP:journals/corr/abs-1811-04407,DBLP:journals/corr/SorokinSPFI15,DBLP:journals/corr/abs-1712-04603}. It is important to consider the spatial and temporal dimensions of the data, and maintain the possibility of attending to different parts of the input across the layers of the network. Our six proposed instantations are described as follows (see Fig.\ref{fig:Self attending network (SAN) architecture}).

\begin{itemize}
	
	\item Self-Attending Network (SAN): Self-attention between convolutional layers 'H1' and 'H2'. This approach focuses on how attention interacts with the input in the lowest level of the network.
	
	\item Strong Self-Attending Network (SSAN): Multiplying the output of the last convolutional layer in the self-attention component ('Y') by a factor of two (thereby increasing the influence of attention on the network).
	
	\item Self-Attending Double Network (SADN): Self-attention between convolutional layers 'H1' and 'H2', 'H2' and 'H3'. Since the higher level layers learn the semantics and higher level abstractions, we intend to evaluate how attention changes the performance when applied to these layers.
	
	\item Strong Self-Attending Double Network (SSADN): Multiplying the outputs for both self-attention components by a factor of two.
	
	\item Pure Self-Attending Network (PSAN): Passing only the output of the self-attention forward, removing the addition of the previous convolutional layer output. This approach investigates the performance of the agent when only the 'pure' sequence representations learnt by the self-attention component are passed forward in the network.
	
	\item Pure Self-Attending Double Network (PSADN): Self-attention between convolutional layers 'H1' and 'H2', 'H2' and 'H3', while passing only the output of the self-attention forward.

\end{itemize}

\subsection{Validation Methodology}

\subsubsection{Implementation}

The Arcade Learning Environment is a well established baseline which allows us to critically evaluate the effects of our proposed architecture modifications. We use Proximal Policy Optimisation \cite{DBLP:journals/corr/SchulmanWDRK17} to train our agents over traditional DQN baselines due to its wall clock training time and improved general performance. In the interest of comparability, the open source implementation from OpenAI 'Baselines'  was utilised \cite{baselines}. In order to objectively identify the effects of the additional attention model, the standards set by Mnih \emph{et al.}~\cite{mnih2015humanlevel} were followed. This included preprocessing of the input image from a single 210x160 RGB image to a stack of four 84x84 grey-scale images. 'No-Op' starts were also used which prevents the agent from taking an action at the start of each game for a random number (maximum thirty) of time-steps.

\subsubsection{Performance evaluation}
In order to evaluate our agents we randomly seed each different architecture for a total of three times across ten different Atari games. In terms of standard training times for bench marking, \cite{mnih2015humanlevel}\cite{DBLP:journals/corr/SchulmanWDRK17}\cite{DBLP:journals/corr/HausknechtS15}\cite{DBLP:journals/corr/abs-1803-00933} show variations between 40M to 16B+ frames.We train each model for a total of 40M time-steps, which is equivalent to 160M frames. This is inline with the evaluation methodology as presented by Fortunato \emph{et al.}~\cite{DBLP:journals/corr/FortunatoAPMOGM17}. Performance is evaluated by the maximal score achieved (after averaging) during training.

\subsubsection{Visual evaluation}
As we are interested in improving an agents ability to reason useful information from its environment, it is important to know what exactly the agent is attending to when making its decisions. The visualisation technique proposed by Greydanus \emph{et al.} involves blurring a part of the input and evaluating the effect on policy performance in order to determine attention at that location. As this would be computationally exhaustive to implement with a sequential series of input frames stacked together, we implement a Grad-cam \cite{DBLP:journals/corr/SelvarajuDVCPB16} inspired approach similiar to Weitkamp \emph{et al.}~\cite{DBLP:journals/corr/abs-1902-0056e}.  We produce action-discriminative activation maps using the gradients back-propagated with respect to the chosen action. Global average pooling is performed over the gradients to determine the neuron importance weights, $ \alpha_{k}^{a} $ of action a, for the last activation layer $k$ in our network. 

\begin{equation}
	\alpha_{k}^{a} =\overbrace{ \frac{1}{Z}\sum_{i}\sum_{j}}^\textrm{global\ average\  pooling}\underbrace{\frac{\partial{h^a}}{\partial{L_{ij}^{K}}}}_\textrm{gradients}
\end{equation}

Where $h^{a}$ is the score for action $a$ prior to the softmax. The gradients are then elementwise multiplied by the forward pass activations of the final activation layer $ L^k $ before passing through a RELU activation, revealing $L^a$, the weighted action activation map.

\begin{equation}
	L^{a} = ReLU(\sum_{k=1}^{K}\alpha_{k}^{a}L^{k})
\end{equation}

This activation map is then bilinearly extrapolated to the size of the input frame and overlaid producing accurate indications of visual attention with respect to decision making. 

\section{Experiments}

\begin{figure*}
	\includegraphics[width=\linewidth]{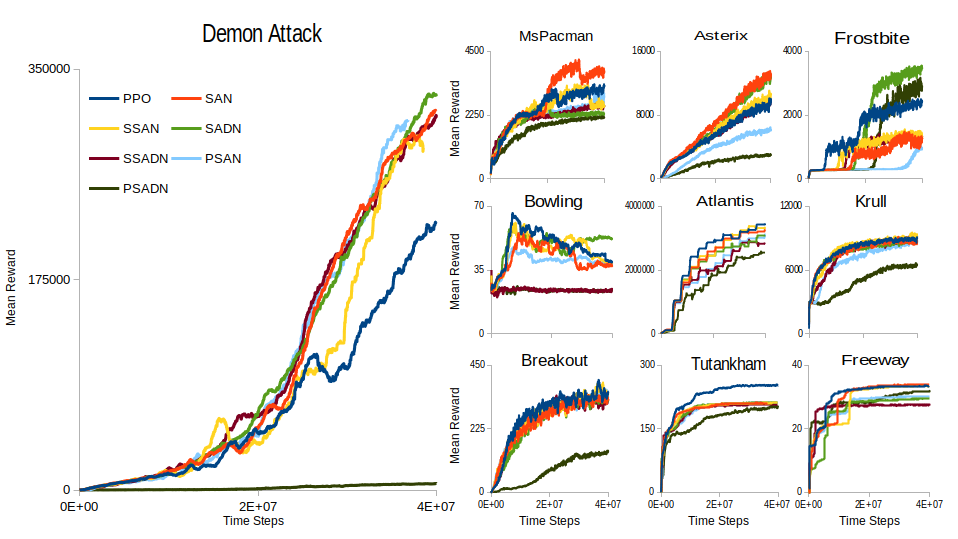}
	\caption{Learning curves of all variants compared to the baseline on all ten environments tested. Agents are trained for a total of 40M timesteps, with results averaged over three random seeds. Here we can see the clear advantage of self-attention is able to provide with respect to sample efficiency.}
	\label{fig:Training_plots}
\end{figure*}

\begin{table}
	\begin{center}
		\begin{tabular}{|l|c|c|c|c|c|c|c|}
			\hline
			& PPO & SAN & SSAN & SADN & SSADN & PSAN & PSADN \\
			\hline
			MsPacman & 3342.15 & \textbf{4217.3} & 3351.4 & 2638.3 & 2681.5 & 3118.6 & 2252.36 \\
			Tutankham & \textbf{254.57} & 211.04 & 211.14 & 212.07 & 208.15 &211.60 &205 \\
			Freeway & 33.48 & \textbf{33.93} & 33.58 & 29.66 & 27.73 & 30.21 & 31.87\\
			Atlantis & \textbf{3445922} & 3272823 & 3387428 & 3093157 & 2958253 & 3120225 & 2583775\\
			Krull & 9102 & 9136 & \textbf{9535} & 8912 & 9191 & 9171 & 6668\\
			Demon Attack & 222650 & 315727 & 294359 & \textbf{329837} & 311438 & 307539 & 5510 \\
			Bowling & \textbf{66.3}3 & 55.21 & 60.99 & 57.52 & 35 & 47.49 & 26.26\\
			Frostbite & 2503 & 1481 & 1552 & \textbf{3554} & 1426 & 1036 & 3173\\
			Asterix & 10121 & \textbf{13556} & 11069 & 13253 & 9815 & 6461 & 3106\\
			Breakout& \textbf{398.10} & 353.47 & 388.89 & 345.07 & 354.81 & 376.45 & 147.81\\
			\hline
		\end{tabular}
	\end{center}
	\caption{Maximal score achieved by each implementation, averaged over three random seeds and trained for 40M time-steps. These results clearly demonstrate the improved performance of multiple self-attention variants.}
	\label{All results}
\end{table}

\subsection{Performance results}

By integrating attention into the underlying neural network, new state of the art results for Demon Attack were achieved. In fact, all but one implementation was able to significantly improve against the baseline, along with surpassing the previous state of the art results for Demon Attack. Additionally, SAN was able to produce significant improvements in both Asterix and MsPacman. Impressively SAN is able to surpass the previously highest score reported using a policy gradient method for Demon Attack, MsPacman, Bowling, Freeway and Frostbite \cite{DBLP:journals/corr/FortunatoAPMOGM17,NIPS2017_7112}.

Table \ref{All results} shows the maximal score after averaging over three random seeds during training for 40M time-steps. From this we can observe that integrating self-attention, in one form or another, led to an increase in performance across 60\% of environments tested. Table \ref{PPO vs SAN} also shows our proposed network, SAN, improved results in 50\% of environments when directly compared to the baseline PPO agent. While Fig. \ref{fig:Training_plots} shows the training curves for each network across all ten environments. This allows us to visually see the increased sample efficiency self-attention provides in envirnoments such as Demon Attack, MsPacman, Asterix and Frostbite. 5

Although single implementations of attention such as SAN and SSAN were able to achieve higher rewards across more environments then other 'double' implementations, it is clear in Fig. \ref{fig:Training_plots} and Table \ref{All results} that SADN is able to outperform SAN in Demon Attack, Frostbite and, Bowling. This provides support for the idea that attention in general is beneficial to the network.

\begin{table}
	\begin{center}
		\begin{tabular}{|c|c|c|}
			\hline
			\textbf{Environment} & \textbf{PPO} & \textbf{SAN} \\
			\hline 
			MsPacman & 3342.15 & \textbf{4217.3} \\
			Tutankham & \textbf{254.57} & 211.04 \\
			Freeway & 33.48 & \textbf{33.93} \\
			Atlantis & \textbf{3445922} & 3272823 \\
			Krull & 9102 & \textbf{9136} \\
			Demon Attack & 222650 & \textbf{315727} \\
			Bowling & \textbf{66.3}3 & 55.21\\
			Frostbite & \textbf{2503} & 1481\\
			Asterix & 10121 & \textbf{13556} \\
			Breakout& \textbf{398.10} & 353.47 \\
			\hline
			
		\end{tabular}	
	\end{center}
	\caption{Direct comparison between baseline PPO and SAN. Clearly demonstrating that our proposed method is capable of improving performance in 50\% of tested environments.}
	\label{PPO vs SAN}
\end{table}

\subsection{Visualisation evaluation}
 
By comparing the visualisation results of the baseline against implementations with self-attention, in particular SAN, a number of insights were produced. These insights include an increased ability to track and attend to multiple targets, a better understading of spatial information from the state input and, temporal attention in situations of partial visibility not seen in the baseline. These situations of partial visibility change the underlying structure of the problem from a Fully Observable Markov Decision Process (FOMDP), to a Partially Observable Markov Decision Process (POMDP). This significantly increase the difficulty of the problem the agent is trying to solve.

\begin{figure}
	\centering
	\subfloat[]{\label{Temporal_attention:a}\includegraphics[scale=.5]{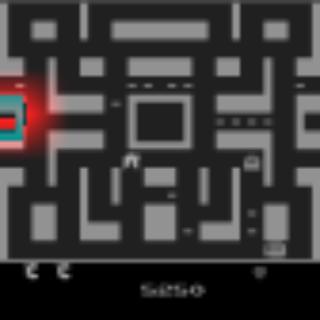}}
	\	\	\ 	\
	\subfloat[]{\label{Temporal_attention:b}\includegraphics[scale=.5]{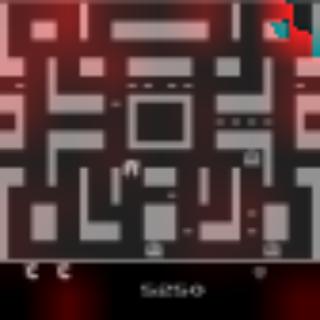}}
	\\
	\subfloat[]{\label{Temporal_attention:c}\includegraphics[scale=.5]{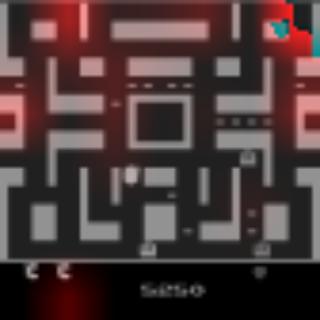}}
	\	\	\	\
	\subfloat[]{\label{Temporal_attention:d}\includegraphics[scale=.5]{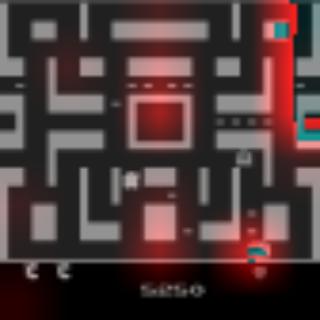}}
	\caption{Here we can see that the agent is able to attend to multiple targets along with using temporal information to compensate for the missing enemy in the most recent frame. The attention also appears to bounce around, as the agent constantly searches the map. As the agent only has temporal information relating to four time steps, this is reasonable behaviour. The hard focus in the top corner is likely related to the fact that the 'super candy' appears in this spot, and like the enemies displays a blinking pattern. As the collecting of this reward has particular benefits to the agent it is not surprising to find the agent attending to this spot. Areas attended to by the agent for decision making are shown by adding the information from the activation map to the red channel of the image. This appears as either red or cyan, and allows for the information in these areas to remain visible to the reader.}
	\label{Temporal_attention}
\end{figure}%

\begin{figure}
	\centering
	\subfloat[]{\label{Tracking:a}\includegraphics[scale=.5]{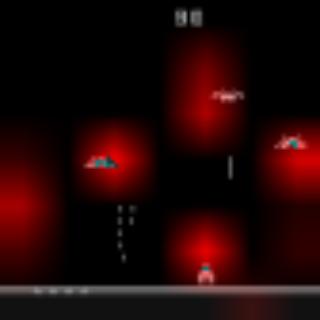}}
	\	\	\ 	\
	\subfloat[]{\label{Tracking:b}\includegraphics[scale=.5]{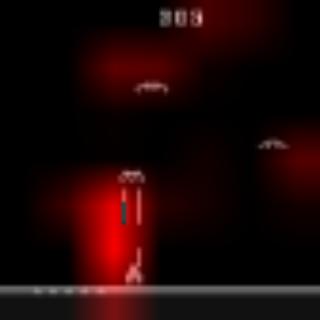}}
	\ 	\ 	\ 	\
	\subfloat[]{\label{Tracking:c}\includegraphics[scale=.5]{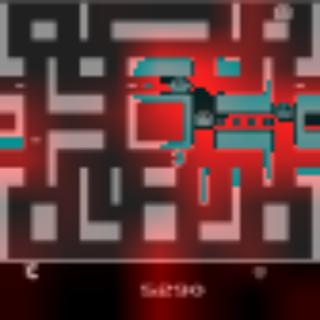}}
	\	\	\ 	\
	\subfloat[]{\label{Tracking:d}\includegraphics[scale=.5]{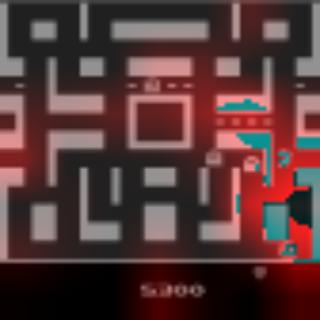}}
	\ 	\ 	\ 	\
	\subfloat[]{\label{Tracking:e}\includegraphics[scale=.5]{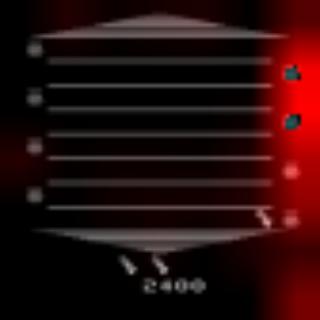}}
	\	\	\ 	\
	\subfloat[]{\label{Tracking:f}\includegraphics[scale=.5]{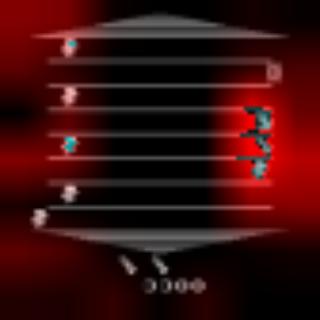}}
	\caption{Examples of a trained SAN agent focusing on multiple enemies in different environments. a,b) Demon Attack. c,d) MsPacman. e,f) Asterix.Areas attended to by the agent for decision making are shown by adding the information from the activation map to the red channel of the image. This appears as either red or cyan, and allows for the information in these areas to remain visible to the reader. }
	\label{Tracking}
\end{figure}

\begin{figure}
	\centering
	\subfloat[]{\label{VP:a}\includegraphics[scale=.5]{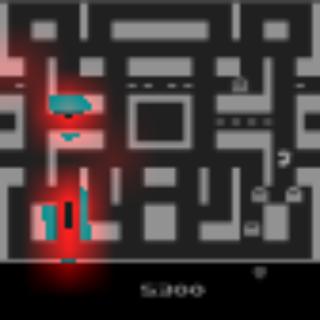}}
	\	\	\ 	\
	\subfloat[]{\label{VP:b}\includegraphics[scale=.5]{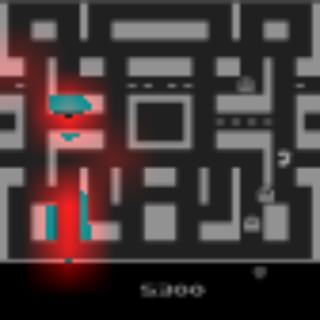}}
	\ 	\ 	\ 	\
	\subfloat[]{\label{VP:c}\includegraphics[scale=.5]{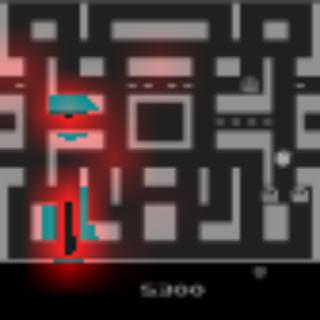}}
	\	\	\ 	\
	\subfloat[]{\label{VP:d}\includegraphics[scale=.5]{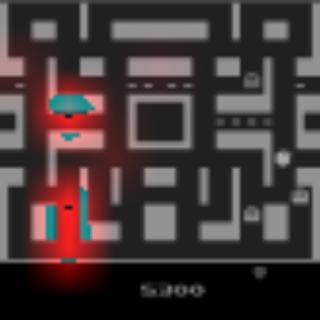}}
	\caption{Examples of disappearing enemies in MsPacman. Enemies are located in the bottom right corner and appear sporadic. This clearly demonstrates the input constitutes a POMDP Areas attended to by the agent for decision making are shown by adding the information from the activation map to the red channel of the image. This appears as either red or cyan, and allows for the information in these areas to remain visible to the reader. }
	\label{VP}
\end{figure}

\begin{figure}
	\centering
	\subfloat[PPO]{\label{main2:a}\includegraphics[scale=.5]{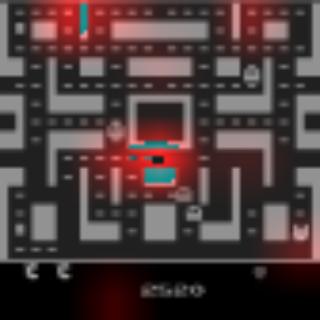}}
	\	\	\ 	\
	\subfloat[SAN]{\label{main2:b}\includegraphics[scale=.5]{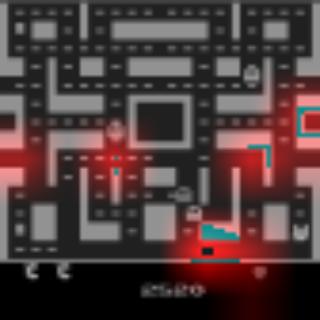}}
	\caption{a) Baseline PPO agent failing to attend to areas of potential interest. b) SAN agent attending to nearby pathways, along with demonstrating spatial awareness by attending to the potential entry point on the other side of the map. This awareness was commonly seen in agents with self-attention. Areas attended to by the agent for decision making are shown by adding the information from the activation map to the red channel of the image. This appears as either red or cyan, and allows for the information in these areas to remain visible to the reader.}
	\label{ PPO-SAN-vis}
\end{figure}%

Fig. \ref{main2:a} shows the baseline PPO agent paying a  very small amount of attention to its surrounding area. While Fig. \ref{main2:b} shows the SAN agent demonstrating attention of its nearby surroundings. It is also clear that the SAN agent is attending to the other side of the map, where enemies can possibly appear from. This level of spatial understanding was common among SAN agents, however was not observed from any of the baseline agents. As our approach has focused on increasing an agents ability to attend to the input information, this difference demonstrates a clear distinction between attention and reason within the agent. The ability to reason with respect to an input is important however, the ability to reason is constrained by the ability to understand the input. The addition of attention to the agent allows it to produce more a meaningful embedding of the state, or put simply, better understand the current input. This ability to better understand the input state allows the agent to make more informed decisions, which is crucial for any reinforcement learning task. Here we show a direct correlation between an agent's ability to understand spatial information and the agent's sample efficiency and performance.

Agents trained with SAN also show a higher level of ability when it comes to tracking multiple enemies or points of interest. Fig. \ref{Tracking} shows the ability of agents trained using SAN to focus on multiple enemies in different environments. This is crucial for any agent to succeed, as was illustrated by Greydanus \emph{et al.} who showed examples of agents failing to attend to nearby enemies resulting in poor performance. 

Referring to Fig. \ref{Temporal_attention}, which shows a sequential time series of four frames, it can be seen that the enemy located in the center bottom of the map is only visible in frames (b) and (c). However in frame (d) the agent attends to the spot were the enemy was last visible. As the agent has all four of the frames as input it is able to see the  enemy in that location, just not on that frame. This is a trait that was only observed in agents trained using self-attention, and is the first visualisations of temporal attention of a RL agent. Importantly the ability to 'look through time' for information is a big advantage in situations of partial observability. 

It should be noted that the missing enemies in MsPacman were a surprise to the authors. No other game tested showed any sign of missing information, yet was a clear and consistent problem in MsPacman. Best practices were followed with all implementations using the standard 'MaxandSkip' environment wrapper, unaltered as supplied from OpenAI 'baselines' \cite{baselines}. This wrapper is designed to ensure the environment conforms as a FOMDP, however it is clear that in this instance the environment is a POMDP. As Atari 2600 games have a tendency to display blinking sprites, due to the memory availabilities of technology at the time, the 'MaxandSkip' environment wrapper essentially compares two frames, seperated by a single time step, and choose the one with the most information. However it would appear that in the case of MsPacman, different enemies are blinking on and off during different frames. This would appear to result in the environment wrapper choosing the best of a bad situation. 

Interestingly MsPacman is one of the few environments left were reinforcement learning has not yielded an agent capable of surpassing the average human level of performance. It is possible that this is due to the POMDP nature of the environment. Observations towards this can be seen in Fig. \ref{VP} where enemies appear to jump sporadically but are in fact different enemies blinking on and off at different frames. We also observed scenarios where enemies disappeared for a total of eight frames prior to the agent making contact, resulting in the death of the agent. As the agent only takes in a history of four frames, it was impossible for the agent, even with temporal attention to avoid this situation.

\section{Conclusions}

We compare the performance of the different implementations against a baseline PPO model. From these results it is clear that the addition of attention is beneficial. Of particular note is the environment Demon Attack, where just about all attention models are able to surpass the previous state of the art results. This was also achieved with a higher sample efficiency compared to other published methods which rely use the traditional network architecture. We can also see clear improvements over baseline results with respect to 60\% of environments tested. We also show, to the best of the authors knowledge the first visualisations of temporal attention with reinforcement learning. This along with the fact that models trained with attention display a higher level of spatial understanding and an increased ability to track multiple points of interest, demonstrates the benefit of combining attention with reinforcement learning.

Future work will seek to further investigate why more attention was beneficial in some environments compared to others, along with further testing of the proposed architecture with different optimisation techniques, including DQN methods.  

\section{Acknowledgments}

We would like to thank Michele Sasdelli for his helpful discussions, and Damien Teney for his feed-back and advice on writing this paper. 


%
%
%
 \bibliographystyle{splncs04}
 \bibliography{egbib}

\begin{thebibliography}{10}
\providecommand{\url}[1]{\texttt{#1}}
\providecommand{\urlprefix}{URL }
\providecommand{\doi}[1]{https://doi.org/#1}

\bibitem{DBLP:journals/corr/BahdanauCB14}
Bahdanau, D., Cho, K., Bengio, Y.: Neural machine translation by jointly
  learning to align and translate. CoRR  \textbf{abs/1409.0473} (2014),
  \url{http://arxiv.org/abs/1409.0473}

\bibitem{bellemare13arcade}
{Bellemare}, M.G., {Naddaf}, Y., {Veness}, J., {Bowling}, M.: The arcade
  learning environment: An evaluation platform for general agents. Journal of
  Artificial Intelligence Research  \textbf{47},  253--279 (jun 2013)

\bibitem{Broadbent_Filter}
Broadbent, D.E.: Perception and communication (1958)

\bibitem{DBLP:journals/corr/abs-1712-04603}
Choi, J., Lee, B., Zhang, B.: Multi-focus attention network for efficient deep
  reinforcement learning. CoRR  \textbf{abs/1712.04603} (2017),
  \url{http://arxiv.org/abs/1712.04603}

\bibitem{baselines}
Dhariwal, P., Hesse, C., Klimov, O., Nichol, A., Plappert, M., Radford, A.,
  Schulman, J., Sidor, S., Wu, Y., Zhokhov, P.: Openai baselines.
  \url{https://github.com/openai/baselines} (2017)

\bibitem{Fang:2018:ALE:3240508.3240571}
Fang, S., Xie, H., Zha, Z.J., Sun, N., Tan, J., Zhang, Y.: Attention and
  language ensemble for scene text recognition with convolutional sequence
  modeling. In: Proceedings of the 26th ACM International Conference on
  Multimedia. pp. 248--256. MM '18, ACM, New York, NY, USA (2018).
  \doi{10.1145/3240508.3240571},
  \url{http://doi.acm.org/10.1145/3240508.3240571}

\bibitem{DBLP:journals/corr/FortunatoAPMOGM17}
Fortunato, M., Azar, M.G., Piot, B., Menick, J., Osband, I., Graves, A., Mnih,
  V., Munos, R., Hassabis, D., Pietquin, O., Blundell, C., Legg, S.: Noisy
  networks for exploration. CoRR  \textbf{abs/1706.10295} (2017),
  \url{http://arxiv.org/abs/1706.10295}

\bibitem{inproceedings}
Gregor, M., Nemec, D., Janota, A., Pirnik, R.: A visual attention operator for
  playing pac-man. pp.~1--6 (05 2018). \doi{10.1109/ELEKTRO.2018.8398308}

\bibitem{DBLP:journals/corr/abs-1711-00138}
Greydanus, S., Koul, A., Dodge, J., Fern, A.: Visualizing and understanding
  atari agents. CoRR  \textbf{abs/1711.00138} (2017),
  \url{http://arxiv.org/abs/1711.00138}

\bibitem{Han:2018:EMR:3265987.3265996}
Han, Y.: Explore multi-step reasoning in video question answering. In:
  Proceedings of the 1st Workshop and Challenge on Comprehensive Video
  Understanding in the Wild. pp.~5--5. CoVieW'18, ACM, New York, NY, USA
  (2018). \doi{10.1145/3265987.3265996},
  \url{http://doi.acm.org/10.1145/3265987.3265996}

\bibitem{DBLP:journals/corr/HausknechtS15}
Hausknecht, M.J., Stone, P.: Deep recurrent q-learning for partially observable
  mdps. CoRR  \textbf{abs/1507.06527} (2015),
  \url{http://arxiv.org/abs/1507.06527}

\bibitem{DBLP:journals/corr/abs-1803-00933}
Horgan, D., Quan, J., Budden, D., Barth{-}Maron, G., Hessel, M., van Hasselt,
  H., Silver, D.: Distributed prioritized experience replay. CoRR
  \textbf{abs/1803.00933} (2018), \url{http://arxiv.org/abs/1803.00933}

\bibitem{DBLP:journals/corr/abs-1802-09070}
Kastaniotis, D., Ntinou, I., Tsourounis, D., Economou, G., Fotopoulos, S.:
  Attention-aware generative adversarial networks (ata-gans). CoRR
  \textbf{abs/1802.09070} (2018), \url{http://arxiv.org/abs/1802.09070}

\bibitem{DBLP:journals/corr/KayCSZHVVGBNSZ17}
Kay, W., Carreira, J., Simonyan, K., Zhang, B., Hillier, C., Vijayanarasimhan,
  S., Viola, F., Green, T., Back, T., Natsev, P., Suleyman, M., Zisserman, A.:
  The kinetics human action video dataset. CoRR  \textbf{abs/1705.06950}
  (2017), \url{http://arxiv.org/abs/1705.06950}

\bibitem{DBLP:journals/corr/MnihHGK14}
Mnih, V., Heess, N., Graves, A., Kavukcuoglu, K.: Recurrent models of visual
  attention. CoRR  \textbf{abs/1406.6247} (2014),
  \url{http://arxiv.org/abs/1406.6247}

\bibitem{mnih2015humanlevel}
Mnih, V., Kavukcuoglu, K., Silver, D., Rusu, A.A., Veness, J., Bellemare, M.G.,
  Graves, A., Riedmiller, M., Fidjeland, A.K., Ostrovski, G., Petersen, S.,
  Beattie, C., Sadik, A., Antonoglou, I., King, H., Kumaran, D., Wierstra, D.,
  Legg, S., Hassabis, D.: Human-level control through deep reinforcement
  learning. Nature  \textbf{518}(7540),  529--533 (Feb 2015),
  \url{http://dx.doi.org/10.1038/nature14236}

\bibitem{DBLP:journals/corr/OhCSL16}
Oh, J., Chockalingam, V., Singh, S.P., Lee, H.: Control of memory, active
  perception, and action in minecraft. CoRR  \textbf{abs/1605.09128} (2016),
  \url{http://arxiv.org/abs/1605.09128}

\bibitem{DBLP:journals/corr/SchulmanWDRK17}
Schulman, J., Wolski, F., Dhariwal, P., Radford, A., Klimov, O.: Proximal
  policy optimization algorithms. CoRR  \textbf{abs/1707.06347} (2017),
  \url{http://arxiv.org/abs/1707.06347}

\bibitem{DBLP:journals/corr/SelvarajuDVCPB16}
Selvaraju, R.R., Das, A., Vedantam, R., Cogswell, M., Parikh, D., Batra, D.:
  Grad-cam: Why did you say that? visual explanations from deep networks via
  gradient-based localization. CoRR  \textbf{abs/1610.02391} (2016),
  \url{http://arxiv.org/abs/1610.02391}

\bibitem{DBLP:journals/corr/abs-1812-01855}
Shi, J., Zhang, H., Li, J.: Explainable and explicit visual reasoning over
  scene graphs. CoRR  \textbf{abs/1812.01855} (2018),
  \url{http://arxiv.org/abs/1812.01855}

\bibitem{DBLP:journals/corr/SigurdssonVWFLG16}
Sigurdsson, G.A., Varol, G., Wang, X., Farhadi, A., Laptev, I., Gupta, A.:
  Hollywood in homes: Crowdsourcing data collection for activity understanding.
  CoRR  \textbf{abs/1604.01753} (2016), \url{http://arxiv.org/abs/1604.01753}

\bibitem{Simonyan2013DeepIC}
Simonyan, K., Vedaldi, A., Zisserman, A.: Deep inside convolutional networks:
  Visualising image classification models and saliency maps. CoRR
  \textbf{abs/1312.6034} (2013)

\bibitem{DBLP:journals/corr/SorokinSPFI15}
Sorokin, I., Seleznev, A., Pavlov, M., Fedorov, A., Ignateva, A.: Deep
  attention recurrent q-network. CoRR  \textbf{abs/1512.01693} (2015),
  \url{http://arxiv.org/abs/1512.01693}

\bibitem{DBLP:journals/corr/abs-1711-07971}
Wang, X., Girshick, R.B., Gupta, A., He, K.: Non-local neural networks. CoRR
  \textbf{abs/1711.07971} (2017), \url{http://arxiv.org/abs/1711.07971}

\bibitem{DBLP:journals/corr/abs-1902-0056e}
Weitkamp, L., van~der Pol, E., Akata, Z.: Visual rationalizations in deep
  reinforcement learning for atari games. CoRR  \textbf{abs/1902.00566} (2019),
  \url{http://arxiv.org/abs/1902.00566}

\bibitem{NIPS2017_7112}
Wu, Y., Mansimov, E., Grosse, R.B., Liao, S., Ba, J.: Scalable trust-region
  method for deep reinforcement learning using kronecker-factored
  approximation. In: Guyon, I., Luxburg, U.V., Bengio, S., Wallach, H., Fergus,
  R., Vishwanathan, S., Garnett, R. (eds.) Advances in Neural Information
  Processing Systems 30, pp. 5279--5288. Curran Associates, Inc. (2017),
  \url{http://papers.nips.cc/paper/7112-scalable-trust-region-method-for-deep-reinforcement-learning-using-kronecker-factored-approximation.pdf}

\bibitem{DBLP:journals/corr/abs-1711-10485}
Xu, T., Zhang, P., Huang, Q., Zhang, H., Gan, Z., Huang, X., He, X.: Attngan:
  Fine-grained text to image generation with attentional generative adversarial
  networks. CoRR  \textbf{abs/1711.10485} (2017),
  \url{http://arxiv.org/abs/1711.10485}

\bibitem{DBLP:journals/corr/abs-1811-04407}
Yuezhang, L., Zhang, R., Ballard, D.H.: An initial attempt of combining visual
  selective attention with deep reinforcement learning. CoRR
  \textbf{abs/1811.04407} (2018), \url{http://arxiv.org/abs/1811.04407}

\bibitem{DBLP:journals/corr/abs-1806-03960}
Zhang, R., Liu, Z., Zhang, L., Whritner, J.A., Muller, K.S., Hayhoe, M.M.,
  Ballard, D.H.: {AGIL:} learning attention from human for visuomotor tasks.
  CoRR  \textbf{abs/1806.03960} (2018), \url{http://arxiv.org/abs/1806.03960}

\bibitem{AAAI1816534}
Zhao, S., Zhang, Z.: Attention-via-attention neural machine translation
  (2018), \url{https://www.aaai.org/ocs/index.php/AAAI/AAAI18/paper/view/16534}

\end{thebibliography}

\end{document}